\documentclass[10pt,twocolumn,letterpaper]{article}

\usepackage{cvpr}              
\usepackage{url}
\usepackage{multicol}
\usepackage{graphicx}
\usepackage{caption}
\usepackage{pifont}
\usepackage{multirow}
\usepackage{amssymb}
\usepackage{booktabs}
\usepackage{array}
\usepackage{bm}
\usepackage{float}
\usepackage{algorithm}
\usepackage{algorithmic}
\usepackage[dvipsnames]{xcolor}

\def\bv{\mathbf{v}}

\def\bn{\mathbf{n}}

\def\bp{\mathbf{p}}

\def\bt{\mathbf{t}}

\def\ba{\mathbf{a}}
\def\bb{\mathbf{b}}

\def\bg{\mathbf{g}}
\def\bS{\mathbf{S}}

\definecolor{cvprblue}{rgb}{0.21,0.49,0.74}
\usepackage[pagebackref,breaklinks,colorlinks,citecolor=cvprblue]{hyperref}

\def\paperID{***} 
\def\confName{CVPR}
\def\confYear{2024}

\vspace{-40pt}
\title{Leveraging Generative Language Models for Weakly Supervised Sentence Component Analysis in Video-Language Joint Learning}

\vspace{-40pt}

\author{
Zaber Ibn Abdul Hakim ${}^{1, *}$
\and
Najibul Haque Sarker ${}^{1, *}$
\and
Rahul Pratap Singh ${}^{2}$
\and
Bishmoy Paul ${}^{1}$
\and
Ali Dabouei ${}^{3}$
\and
Min Xu ${}^{3}$ \\
\vspace{-13pt}
\and
${}^{1}$Bangladesh University of Engineering and Technology
\and
${}^{2}$Netaji Subhas University of Technology
\and
${}^{3}$Carnegie Mellon University
}

\begin{document}
\maketitle
\def\thefootnote{*}\footnotetext{Equal Contribution}\def\thefootnote{\arabic{footnote}}
\begin{abstract}
\vspace{-10pt}
A thorough comprehension of textual data is a fundamental element in multi-modal video analysis tasks. However, recent works have shown that the current models do not achieve a comprehensive understanding of the textual data during the training for the target downstream tasks.  
Orthogonal to the previous approaches to this limitation, we postulate that understanding the significance of the sentence components according to the target task can potentially enhance the performance of the models. Hence, we utilize the knowledge of a pre-trained large language model (LLM) to generate text samples from the original ones, targeting specific sentence components. We propose a weakly supervised importance estimation module to compute the relative importance of the components and utilize them to improve different video-language tasks. Through rigorous quantitative analysis, our proposed method exhibits significant improvement across several video-language tasks. In particular, our approach notably enhances video-text retrieval by a relative improvement of 8.3\% in video-to-text and 1.4\% in text-to-video retrieval over the baselines, in terms of R@1. Additionally, in video moment retrieval,  average mAP shows a relative improvement ranging from 2.0\% to 13.7 \% across different baselines.

\end{abstract}
\vspace{-5pt} 
\vspace{-10pt}
\section{Introduction}
\label{sec:intro}
\vspace{-7pt}

\begin{figure}
  \centering
    \includegraphics[scale=1.2]{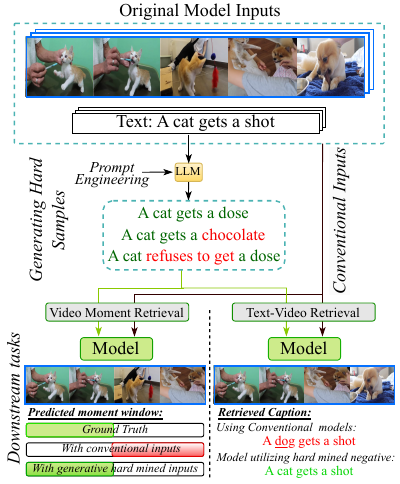}
    \caption{Performance comparison between existing approaches and our proposed method in video-language joint tasks. Here, conventional models fail to properly attend all parts of the sentence, \eg \textbf{\textit{shot}}. This is alleviated by the incorporation of targeted hard samples using our proposed mechanism.}
    \label{fig:umt_error}
    \vspace{-20pt}
\end{figure}

With the rise of social media, streaming platforms, surveillance systems, and the entertainment industry, video has been solidified as one of the prime sources of information and recreation. Natural language is one of the most important modalities that accompanies video data due to its human-friendly and descriptive nature. Inclusion of both modalities in a unified task requires the interaction between videos and user input texts to interact in a multi-modal domain, giving rise to video-language joint learning tasks that include Video Retrieval \cite{xpool, ma2022x, dzabraev2021mdmmt}, Video Captioning \cite{iashin2020multi, zhang2020object}, Action Segmentation \cite{farha2019ms, chen2020action}, Video Moment Retrieval \cite{qddetr, lei2021detecting, umt}, Video Summarization \cite{gao2020listen}, etc. The performance of models in these tasks depends on their capability to extract video features and align them with text features. Attending equally to each part of a sentence and assigning them the appropriate level of significance with regard to the video modality is an important requirement of these models. Our exploratory analysis suggests that the current models do not attend to all parts equally which results in incorrect outputs. One such instance is shown in Figure \ref{fig:umt_error}.
\begin{table*}[t]
\vspace{-5pt}
    \small
    \centering 
    \begin{tabular}{|c|c|c|c|} 
        \hline
        \textbf{Anchor Text (A)}  & \textbf{type} & \textbf{Generated Samples (M)} & \textbf{\begin{tabular}[c]{@{}c@{}}Similarity \\  between A \& M \end{tabular}}  \\
        \hline

        \multirow{2}{*}{{\begin{tabular}[c]{@{}c@{}}Young tourist couple sharing \\ some videos of their \underline{tour}. \end{tabular}}}
          &  Negative  & {\begin{tabular}[c]{@{}c@{}}Young tourist couple sharing some 
 \\  videos of their \textcolor{red}{wedding}. \end{tabular}}  & 0.906\\

        \cline{2-4} 

         &  Positive & {\begin{tabular}[c]{@{}c@{}}Some videos of their tour are being 
 \\  shared by the young tourist couple. \end{tabular}}   & 0.881\\

    \hline
        
    \end{tabular}
    \caption{Example of negative and positive samples generated using LLMs. By utilizing the capabilities of the LLMs we can generate very hard samples. It is supported by the generated negative samples being much closer to anchor text in embedding space compared to the corresponding positive text.}
    \label{tab:example}
    \vspace{-15pt}
\end{table*}
Here in moment retrieval, the baseline model shows bias towards the text subject \emph{`cat'} and fails to connect it with the object \emph{`shot'}. A parallel situation occurs in video-to-text retrieval, where the model struggles to distinguish that the subject \emph{`dog'} in the retrieved text is unrelated to the object \emph{`shot'}. Consequently, although both the models' predictions include the subject, the object is absent. This problem stems from the models' difficulty in accurately representing all sentence components while aligning with extracted video features.

The limited perception of textual features in video-language joint learning tasks can be rooted back to the presence of noisy text labels in web-crawled image-caption pairs \cite{lai2023scarcity, fan2023improving} used in the pre-training stage of encoders such as CLIP \cite{Radford2021LearningTV}. Correcting the huge amount of text labels in the pre-training stage is a significant challenge, motivating numerous efforts \cite{yang2021taco, he2021improving, falcon2022learning, cap4video, li2023distilling} to improve the textual representation in many downstream tasks with limited dataset.  A common strategy is to introduce extra weak labels, aiding models in distinguishing sentence differences and ultimately enhancing feature representation. In earlier works \cite{yang2021taco, he2021improving, zolfaghari2021crossclr}, such additional negative and positive samples have been mined from the original dataset based on the similarity of the text representations. While the results are promising, the procedure for generating additional samples presented in those works is not controllable and there is no way to emphasize specific sentence parts.

A popular solution among researchers to generate additional labels in a more controllable manner involves leveraging the vast knowledge of a pre-trained large language model (LLM). In the majority of the cases \cite{cap4video, li2023distilling, lai2023scarcity, fan2023improving}, LLM has been used to generate descriptive and refined information from the videos or the text labels. Apart from this, in some concurrent works \cite{doveh2023teaching, momeni2023verbs}, LLM has also been used to introduce completely new positive or negative samples to be used in a contrastive learning setting. Although \cite{momeni2023verbs} emphasized the significance of verbs in video-language joint learning, a comprehensive investigation into various sentence components, such as objects, subjects, adjectives, etc., and their relative importance in understanding video-text correlation remains unexplored. In this work, we leverage LLM to generate hard negative samples from the original (anchor) samples that emphasize different sentence parts. This involves using precise prompt engineering to modify specific parts of the sentence while keeping the rest unchanged. Additionally, by completely restructuring the sentence, we create positive samples that lie relatively far from the anchor in the embedding space, compared to their negative counterparts, as shown in Table \ref{tab:example}. 

We incorporate these generated samples using a modified contrastive loss function that considers the relative importance of different sentence parts. A visual comparison between an existing approach and our approach has been illustrated in Figure \ref{fig:umt_error}.


To sum up, the main contributions of our work are:

\begin{itemize}
    \item We devise a mechanism for generating hard negative and positive samples for video-text joint learning tasks that emphasize different sentence parts. 
    \item We propose a pipeline that utilizes the generated samples to evaluate the importance of different sentence components for the computation of adaptive contrastive loss.
    \item Through extensive quantitative evaluations on two major video-text joint learning tasks, we demonstrate consistent improvement of performance from the baseline in both scenarios. Furthermore, we conduct qualitative investigations to provide insights into the models' decision-making process after the integration of our approach.
\end{itemize}

\section{Related Works}\label{sec:related_works}

\begin{figure*}[t]
  \centering 
  \vspace{-25pt}
 \includegraphics[scale=0.3]{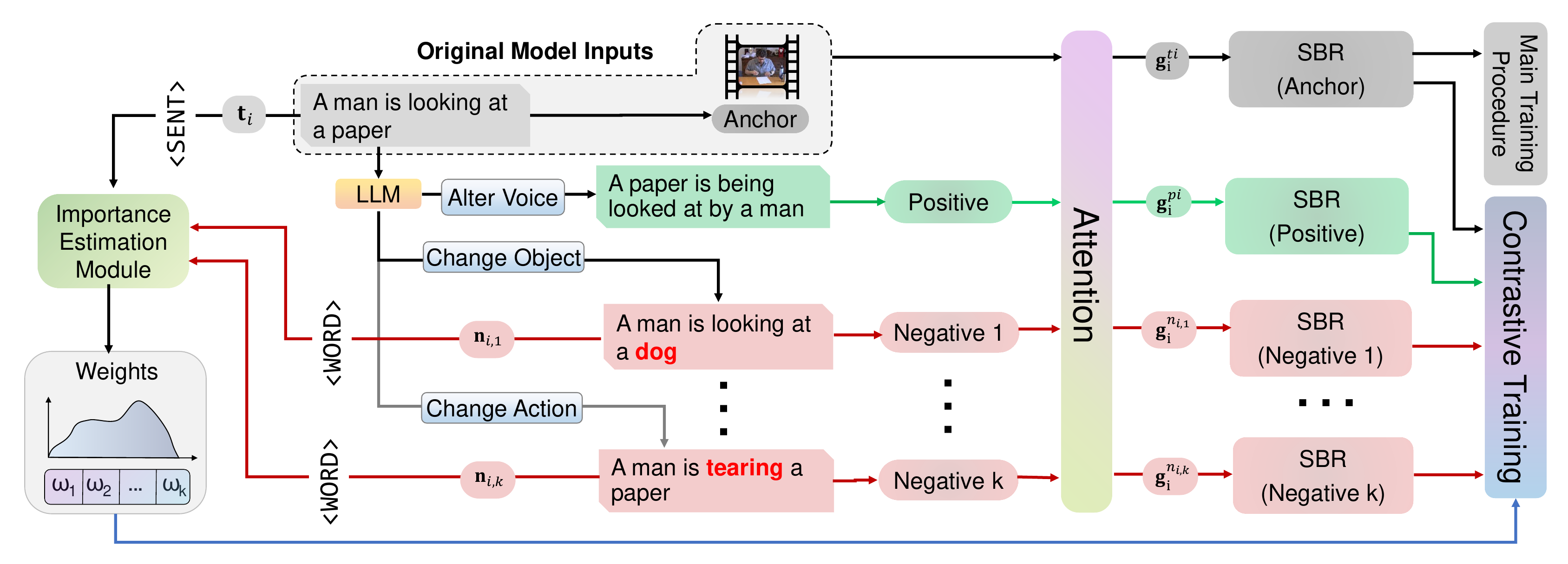}
 \vspace{-8pt}
   \caption{An overview of our proposed method. The videos, original or anchor texts, generated positive texts, and negative texts are passed through the Attention mechanism to generate  Similarity-Based Representations, denoted as SBR in the figure (The exact method of generation varies with different tasks). The Contrastive Training procedure utilizes the generated texts to improve video-language context understanding. The Importance Estimation Module uses both the sentence and word representations (denoted as \texttt{<SENT>} and \texttt{<WORD>} respectively) to assign a weight to the losses associated with each of the negative samples.}
   \label{fig:general_figure}
   \vspace{-15pt}
\end{figure*}

Multi-modal video-language joint learning tasks such as moment retrieval and video retrieval aim to establish meaningful alignment between the embedding spaces of textual and visual modalities \cite{vatex, DisentangledRL}. 

Recent advancements in the field have leveraged pre-trained encoders, \eg, CLIP \cite{Radford2021LearningTV}, as the feature extractor. They also use contrastive loss to align global or fine-grained representations of the text and video embeddings. 
Models utilizing global representations \cite{ma2022x, xpool} typically focus on \texttt{[CLS]} token for text and joint representation of frame-level embeddings. In contrast, models employing fine-grained representations \cite{10.1145/3397271.3401075, qvh, qddetr} delve into the alignment of word-level representation of text with frame-level representations of video.

\subsection{Contrastive Learning}
Contrastive learning is based on minimizing the distance between similar semantic features (positive pairs) while maximizing the distance between dissimilar ones (negative pairs) \cite{contrastive_survey}. 
In moment retrieval, \cite{ma2020vlanet} utilizes contrastive loss where video-text pairs are positive samples if they belong in the training data. To align video and textual representations, \cite{nan2021interventional} uses target moment clips as positive examples and other clips in that video as negative examples in a dual contrastive setting. 
While \cite{wang2021visual} mines hard positive video moments using query similarity across different videos, \cite{zheng2022weakly} generates easy and hard negative moments from the same video using a learnable Gaussian mask.

\vspace{-6pt}
\subsection{Additional Sample Generation}
\vspace{-4pt}
The usage of additional text samples, specifically hard-negative samples, has been explored in various literature \cite{robinson2020contrastive, wang2021instance, cao2022exploring}. Existing methods of hard-negative sampling are based on extracted features or using metadata and supervision labels \cite{max_sampling, nearest_neighbor_sampling, supervised_hard_neg, jiang2019svd}. In the context of multi-modal tasks, \cite{yang2021taco} uses a cascading sampling method to mine video-text pairs as negative samples in a batch, in contrast to the random sampling done by \cite{luo2020univl, zhu2020actbert}.

\cite{he2021improving} utilizes the adaptive margin mechanism \cite{semedo2019cross} for video retrieval as a better alternative to random sampling. \cite{falcon2022learning} introduces Relevance-Aware Negatives and Positives mining (RANP) which uses semantics to mine samples. On the other hand, \cite{zolfaghari2021crossclr} excludes strongly connected negative samples by using embeddings to select better negative samples. \cite{xu2021videoclip} generates hard negatives through nearest neighbor retrieval. 
Recently the potential of Large Language Models (LLMs) has been explored in additional sample mining. Few works \cite{li2023distilling, cap4video} explored LLMs by generating additional comprehensive details from the available modalities, such as image captions or videos.
\cite{doveh2023teaching} generates positive samples via LLM prompt engineering and hard negative samples by first masking sentence parts and then unmasking using LLM.

\section{Methodology}
\label{sec:methodology}

To address the existing models' limitations in correlating major sentence parts with suitable video representations, we present a method for generating challenging negative and positive samples targeting specific sentence parts. These samples facilitate improved perception of specific parts of the sentence, eventually increasing the understanding of video-language correlation. We use the generated samples as auxiliary samples alongside the original training samples by employing a novel adaptive contrastive loss. The proposed approach is application-agnostic and can be adopted successfully in any video-language task. Our approach is summarized in Figure \ref{fig:general_figure} 

This section is organized as follows:
We outline our sample generation procedure in Subsection \ref{subsec:sample_gen}. Followed by this, Subsection \ref{subsec:cons_loss} describes the procedure of applying our proposed contrastive loss by incorporating generated samples. In Subsection \ref{subsec:weight_module}, we introduce our importance estimation module that adaptively weighs different sentence components based on their saliency. Finally, Subsection \ref{subsec:va_attn} provides the details of different video-language tasks on which we evaluate our method.

\subsection{Sample Generation}
\label{subsec:sample_gen}
Let $t_1, \dots, t_m$ be the text samples available in a video-language joint learning task, where $m$ is the total number of text samples in the training set. This textual information is the input to the conventional models and as we discussed previously, a major shortcoming of these models is that they do not attend to all information in the sentence well enough. To force the models to have a better understanding of the correlation of different sentence parts with videos, we generate hard negative and positive samples focusing on these components which are later employed in the training. We use a pre-trained LLM to generate such samples by utilizing their huge prior knowledge of the language.

\noindent
\textbf{Generative hard Negative and Positive Sample Mining.} Given the anchor text $t_i$, we generate $k$ hard negative texts, $n_{i,1}, n_{i,2}, \dots, n_{i,k}$, and a positive text, $p_i$, by leveraging the linguistic capability of a pre-trained LLM.  For the generation of negatives, we instruct the LLM to specifically change a sentence part, \ie verb, object, subject, etc. For the majority of the cases, the LLM only modifies the targeted part of the anchor text, keeping the rest of the sentence the same. Conversely, when generating the positive sample, we instruct the LLM to generate a sample that has a completely different sentence structure from the anchor while maintaining its semantics. 

We formalize the process of generating a negative and positive sample as: 
\begin{equation}
\small
    \begin{split}
    n_{i,j} ~\leftarrow ~\texttt{LLM} ( t_i, \text{``Change $<part>_j$ of the sentence"} ),\\
    p_{i} ~\leftarrow~ \texttt{LLM} ( t_i, \text{``Alter voice of the sentence"} ), ~~~~~~~~&
    \end{split}
\end{equation}
where $i \in \{ 1, \dots, m\}$, $j \in \{ 1,\dots, k\}$, and the variable $``part"$ represents any of the $k$ sentence parts which are modified to generate the negative sample.
Finally, we use CLIP's \cite{radford2021learning} text encoder to generate text embeddings from different types of texts. In our formulation, we have used similar notations ( $\bt, \bn, \bp$) to denote both texts and text embeddings interchangeably.

\subsection{Incorporating the Generated Samples}
\label{subsec:cons_loss}
The generated hard negative and positive samples force the existing models to discern the distinction among different words for a specific sentence part and their association with the video. This improves the overall perception of the video-language correlation. To incorporate these additional samples, along with the model's original samples, we compute a contrastive loss from the new samples and use it in association with the model's original loss. We use the embeddings of the three types (anchors, negatives, and positives) of texts mentioned in the preceding subsection, or their composite embedding with the videos to compute the contrastive loss. This approach facilitates the effective utilization of the generated additional auxiliary samples.

Let $\bg_i$ be a general embedding for $i^{th}$ sample, which can either be text embeddings or video-text composite embeddings. Given three types of text embeddings ($\bt_i, \bn_{i,j}, {\bp_i}$), the general embedding for anchor texts, negative texts, and positive texts are denoted with $\bg_i^{t_i}, \bg_i^{n_{i,j}}$, and $\bg_i^{p_i}$ correspondingly. Then the contrastive loss for $i^{th}$ sample is formulated as following:
\begin{equation}
\small
    \mathcal{L}_i = - \log \frac{ e^{ \text{sim} ( \bg_i^{t_i}, \bg_i^{p_i} ) / \tau}}{ e^{ \text{sim} ( \bg_i^{t_i}, \bg_i^{p_i} ) / \tau} + \sum_{\bn \in \mathbf{S}_i} e^{ \text{sim} ( \bg_i^{t_i}, \bg_i^{n} ) / \tau} },
    \label{eqn:cons_act}
\end{equation}
where $\tau$ is the temperature coefficient, $\mathbf{S}_i$ is the set of embeddings of all negative texts for $i^{th}$ sample, and $\text{sim}(\cdot, \cdot)$ represents the function to compute the similarity between the two embeddings.

\begin{figure}[t]
    \includegraphics[scale=0.26]{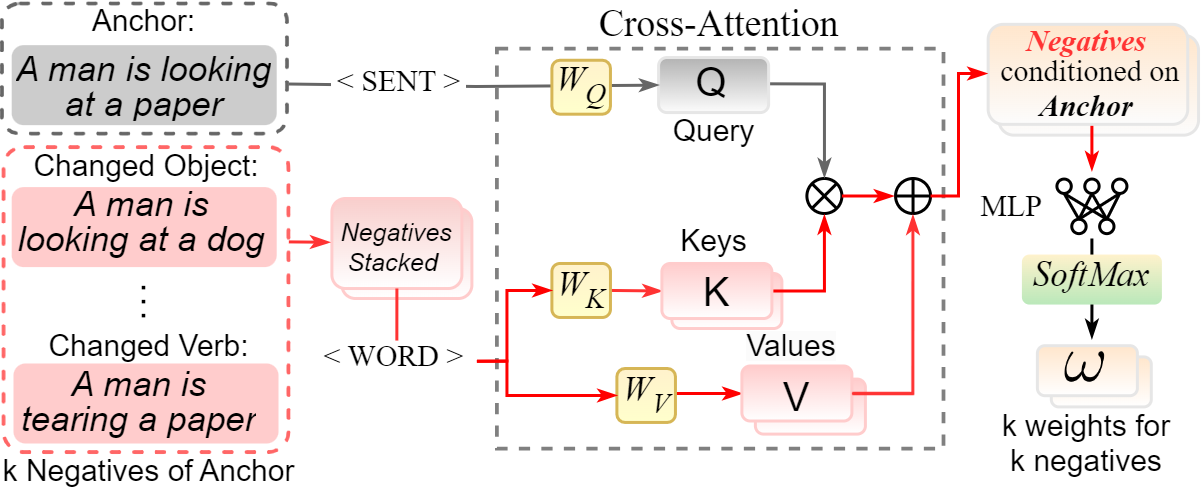}
    \vspace{-5pt}
   \caption{Importance Estimation Module. The sentence and word representations (denoted as \texttt{<SENT>} and \texttt{<WORD>} respectively) from Anchor and Negative text are passed through the cross-attention module to generate weights. The $\bigotimes$ operation denotes the dot products of the keys with the query followed by a softmax operation, and the \scalebox{1.3}{$ \bm{\oplus} $} operation is the weighted averaging of all word-level tokens of values.}
   \label{fig:weight_generation}
   \vspace{-15pt}
\end{figure}

\subsection{Weakly Supervised Importance Estimation of Sentence Components}
\label{subsec:weight_module}

Although we generate multiple types of negative texts by modifying specific components, \eg noun, verb, or object, different sentence components don't exert similar importance in context understanding. For example, objects can play a more important role in some sentences whereas in a different example, it might be irrelevant. The aforementioned concern makes it important to evaluate the relative importance of each type of component adaptively for each sample. In the contrastive loss, defined in Equation \ref{eqn:cons_act}, all the negatives are treated similarly. Considering this, we adaptively choose the most discernable sentence component per instance instead of using all of them together. The positive effect of this strategy highlights the importance of finding an optimal combination of all sentence components. To address this, we introduce a weakly-supervised sentence component analysis module. This module adaptively predicts the saliency of sentence components for each anchor text without any direct supervision. These dynamically computed scores denote which sentence components are crucial in context understanding, as depicted in Figure \ref{fig:result}.

\noindent\textbf{Using the Most Discernable Component.} As we are using completely unsupervised sample generation, some generated samples might be completely unrelated to the anchor sentence. For instance, when a sentence doesn't have any adjective, using an adjective-changed negative won't make much sense. To address this issue, we focus only on the most discernible component. 

To achieve this, we decompose the general contrastive loss of the $i^{th}$ sample, $\mathcal{L}_i$, into $k$ contrastive losses ($\mathcal{L}_{i,1}, \dots, \mathcal{L}_{i,k}$), each corresponding to a specific negative. We compute them with a slight modification to the formulation presented in Equation \ref{eqn:cons_act}. Instead of using a single set of negative text embeddings ($\mathbf{S}_i = \{ \bn_{i, 1}, \dots \bn_{i, k} \}$), we compute the losses using $k$ different sets of negatives, each containing one type of negative sample. The minimum of these decomposed losses is related to the component that the model can identify most confidently. So, we compute the loss for $i^{th}$ sample as below: 
\begin{equation}
    \mathcal{L}_i = \min ( \mathcal{L}_{i, 1}, \mathcal{L}_{i, 2}, \dots, \mathcal{L}_{i, k} )
\end{equation}

\noindent
\textbf{Computation and utilization of Importance Weight Estimation.}
This module utilizes a cross-attention mechanism to attend each type of negative text embedding with the anchor text embedding. We aim to consider the association between the sentence representation of anchor text embedding, $\mathbf{st}_i$, with all of its corresponding word-level representations of negative texts, $\mathbf{n}_{(i, :)}$, to generate a weight for each of the negatives. After applying the attention, we use a linear transformation followed by a softmax activation to generate the weights. 

Let $dim$ be the vector dimension of each token in the text embeddings. $W_Q, W_K, W_V \in \mathbb{R}^{dim \times h}$ denote the learnable weight vectors for the linear transformations to transform the text representations into ``query", ``key", and ``value" correspondingly, where $h$ is the dimension of the hidden state. In addition, $W_{\omega} \in \mathbb{R}^{h}$ is the weight vector for the final linear transformation layer that is applied on the weighted ``value" vector. The operations are formulated as:
\vspace{-1pt}
\begin{equation}
\small
    \begin{split}
        Q_i &= W_Q * \mathbf{st}_{i}, 
        K_{(i, :)} = W_K * \mathbf{n}_{(i, :)}, 
        V_{(i, :)} = W_V * \mathbf{n}_{(i, :)}, \\
        &~~~V'_{(i, :)} = \texttt{Cross-Attention} \left(Q_i, K_{(i, :)}, V_{(i, :)} \right),\\
        &~~~~~~~~~~~~~~~~~~~~~~~~\mathbf{m}_{(i, :)} = W_{\omega} * V'_{(i, :)},\\
        &~~~~~~~~~~~~~~~~~~~~~~~~\bm{\omega}_{(i, :)} = \frac{e^{\mathbf{m}_{(i, :)}}}{\sum_{j} e^{\mathbf{m}_{(i, j)}}},\\
    \end{split}
\end{equation}
where $\bm{\omega}_{(i, :)} \in \mathbb{R}^{k}$ the estimated importance scores. 

Finally, rather than using the simple contrastive loss, illustrated in Equation \ref{eqn:cons_act}, we combine the component-wise contrastive losses, $\mathcal{L}_{i,1}, \dots, \mathcal{L}_{i, k}$,  as below:
\vspace{-5pt}
\begin{equation}
    \mathcal{L}_i = \sum_{j=1}^{k} \bm{\omega}_{i,j} * \mathcal{L}_{i,j}.
\end{equation}
\vspace{-9pt}


\subsection{Attention Between Video and Text}
\label{subsec:va_attn}
For generating the multi-modal feature in video-language tasks, different baseline works adopted various approaches which is also reflected in the general definition of similarity computation in Equation \ref{eqn:cons_act}. In a subset of works \cite{ma2022x, ge2022bridging, croitoru2021teachtext}, simple cosine similarity between video and text embeddings is employed for such purpose. On the other hand, the majority of the works tend to use different variations of attention mechanisms \cite{lei2021detecting, qddetr, liu2021hit}. Given this diversity, there is no fixed approach for all video-language joint learning tasks to generate this multi-modal feature. In this subsection, we discuss such differences and highlight how we adopt different baseline works into our approach. 
\vspace{-7pt}
\subsubsection{Video Moment Retrieval}
Generally, video moment retrieval works with video embeddings that have been attended by text embeddings. To achieve this, a multi-head self-attention layer \cite{vaswani2017attention, wu2019neural} or a cross-modal attention layer \cite{badamdorj2021joint, ye2019cross, wei2020multi} has been used in recent works \cite{qddetr, lei2021detecting, liu2022umt}. The self-attention approach concatenates the video and text embeddings on sequence dimension and then passes it to the conventional self-attention layer \cite{lei2021detecting}. Conversely, in the case of cross-modal attention, videos are used as queries and texts are treated as keys and values for the attention mechanism\cite{qddetr, liu2022umt}.

We formalize a generalizable formulation for the aforementioned attention mechanism that generates the joint embedding, $\bg_i$,  used in Equation \ref{eqn:cons_act}  as below: 
\begin{equation}
\small
    \begin{aligned}
        \bg_i^{a_i} &= \texttt{Attention}_{v-a}(\bv_i, \ba_{i})
    \end{aligned}
\end{equation}
where $\bv_i$ is the video embedding, $\ba_i$ can be the text embedding of either anchor, negative, or positive texts of $i^{th}$ sample. Then the $\texttt{sim}(\cdot, \cdot)$ function in Equation \ref{eqn:cons_act} is represented as a simple cosine similarity of the aforementioned joint embeddings ($\bg_i^{a_i}$).

\vspace{-7pt}
\subsubsection{Video-Text Retrieval}
Video-text retrieval tasks generally work by aligning the videos and their corresponding texts in a shared embedding space. While, initial studies \cite{LUO2022293, xpool} typically focused on the global representations of texts and videos for computing similarity score, recent studies \cite{10.1145/3397271.3401075, yao2022filip, ma2022x} have shown significant improvement by employing word-level and frame-level embeddings in addition to global representations. This enables better alignment of video frames with text words.

We formulate the procedure that generates the similarity score between any text pair as follows:
\begin{equation}
\small
    \begin{aligned}
        S_{t-a} &= \texttt{Attention}_{t-a}(\bt_{i}, \ba_{i}),
    \end{aligned}
    \label{eqn:Attention_t-t}
\end{equation}
where $\bt_{i}$ is the word-level embedding of any anchor text, and $\ba_i$ represents the same for any positive or negative texts of $i^{th}$ sample. The details relating to the implementation of Function \ref{eqn:Attention_t-t} are elaborated in Section \ref{sec:AttentionBasedAggregation} of the supplementary material. This function takes the place of the $\texttt{sim}(\cdot, \cdot)$ function in contrastive loss Equation \ref{eqn:cons_act}.

\begin{table*}[t!]
\vspace{-10pt}
\begin{tabular}{c|cc|ccc|c|cc|ccc}
\hline \noalign{\vskip 2pt}
 & \multicolumn{5}{c|}{Moment-DETR \cite{qvh}} & & \multicolumn{5}{c}{QD-DETR \cite{qddetr}}\\ \cline{2-6} \cline{8-12}
 & \multicolumn{2}{c|}{R1} & \multicolumn{3}{c|}{mAP} & & \multicolumn{2}{c|}{R1} & \multicolumn{3}{c}{mAP}\\ 
 & \multicolumn{1}{c}{@0.5} & @0.7 & \multicolumn{1}{c}{@0.5} & \multicolumn{1}{c}{@0.75} & Avg & & \multicolumn{1}{c}{@0.5} & @0.7 & \multicolumn{1}{c}{@0.5} & \multicolumn{1}{c}{@0.75} & Avg\\ \noalign{\vskip 2pt} \hline 
\hline \noalign{\vskip 2pt}
Baseline & \multicolumn{1}{c}{52.89} & 33.02 & \multicolumn{1}{c}{54.82} & \multicolumn{1}{c}{29.40} & 30.73 & & \multicolumn{1}{c}{62.40} & 44.98 & \multicolumn{1}{c}{62.52} & \multicolumn{1}{c}{39.88} & 39.86\\ 

Simple Contrastive Loss  & \multicolumn{1}{c}{55.32} & 36.45 & \multicolumn{1}{c}{56.88} & \multicolumn{1}{c}{32.88} & 33.75 & & \multicolumn{1}{c}{61.80} & 45.01 & \multicolumn{1}{c}{62.23} & \multicolumn{1}{c}{40.39} & 40.58  \\ 

Most Discernable & \multicolumn{1}{c}{56.61} & 36.71 & \multicolumn{1}{c}{57.64} & \multicolumn{1}{c}{33.83} & 34.26 & & \multicolumn{1}{c}{62.19} & 44.94 & \multicolumn{1}{c}{\textbf{62.62}} & \multicolumn{1}{c}{40.49} & 40.34  \\


Adaptive & \multicolumn{1}{c}{\textbf{56.81}} & \textbf{37.42} & \multicolumn{1}{c}{\textbf{58.30}} & \multicolumn{1}{c}{\textbf{33.95}} & \textbf{34.94} & & \multicolumn{1}{c}{\textbf{63.04}} & \textbf{45.85} & \multicolumn{1}{c}{62.59} & \multicolumn{1}{c}{\textbf{41.29}} & \textbf{40.66} 

\\ \noalign{\vskip 2pt}
\hline
\end{tabular}%
\vspace{-5pt}
\caption{Performance comparison of different settings on QVHighlights test set for moment retrieval. In the experiments other than baseline, all the negative samples have been used together.}
\label{tab:moment_scores}
\vspace{-8pt}
\end{table*}

\section{Exeriments}

\textbf{Datasets.} We use the QVHighlights \cite{lei2021detecting} dataset for performance evaluation on moment retrieval task. It contains over $10{,}000$ videos and each video includes relevant clips associated with human-written text queries. Additionally, we use the MSVD \cite{MSVD} dataset to evaluate performance on video and text retrieval tasks. It contains $1{,}970$ videos and $80$K captions, averaging ${\sim}40$ captions per video. We use the split proposed by Xu et al. \cite{MSVD_split}.

\noindent
\textbf{Implementation Details.} We experiment using $2$ NVIDIA RTX $3090$ $24$ GB GPUs using the Pytorch library. We use the default implementation of Moment-DETR \cite{qvh} and QD-DETR \cite{qddetr} for moment retrieval and X-CLIP for video-text retrieval tasks. We train X-CLIP on a batch size of $30$ with $6.1e^{-5}$ as the learning rate. Further details of our setup are provided in the Supplementary Material Section \ref{sec:eval_protocols}.

\vspace{-3pt}
\subsection{Performance Comparison}
\vspace{-2pt}
\subsubsection{Moment Retrieval}
\vspace{-2pt}
\textbf{Moment-DETR \cite{qvh}.} The results on the test set of QVHighlights for the Moment-DETR model are presented in Table \ref{tab:moment_scores}. Here, we consider the non-pretrained version of the Moment-DETR model as the baseline.  As conventional models are deficient in attending to different parts of the sentences relevant to the video, we utilize prompt engineering to create hard negative samples by targeting individual components of the sentence: verb, adjective/adverb, subject, and object. Additionally, we also convert the texts to negated passive, which generates samples with accentuated objects. Initially, we experiment with a simple contrastive loss configuration on positive and hard negative samples generated by LLM. This increases the model's performance from the baseline's average mAP score of $30.73$ to $33.75$, which is a relative improvement of $9.8\%$, suggesting that by using targeted hard negatives, the model learns to attend to all parts of the sentence.

Subsequently, we devise a contrastive loss setting where instead of using all the generated negatives for a sample, we choose the one that is most discernable for that specific sample. The objective is to observe how the performance is affected when negatives are selected based on the input sample. This further outperforms the baseline average mAP by $11.48\%$ and validates our assumption that for each sample text, there is an optimal combination of negatives which enables the model to put proper emphasis on the relevant part of the sentence. Consequently, after introducing our proposed methodology of weakly supervised adaptive importance estimation of sentence components, the score further improves to $34.94$ for average mAP outperforming the baseline Moment-DETR model by $13.7\%$.

\begin{table}[t]
\resizebox{\columnwidth}{!}{%
\small
\begin{tabular}{c|ccc|ccc}
\hline \noalign{\vskip 2pt}
 & \multicolumn{3}{c|}{T2V} & \multicolumn{3}{c}{V2T} \\ 
\textbf{} & \multicolumn{1}{c}{R@1$\uparrow$} & R@5$\uparrow$ & MnR$\downarrow$ & \multicolumn{1}{c}{R@1$\uparrow$} & \multicolumn{1}{c}{R@5$\uparrow$} & MnR$\downarrow$ \\  \noalign{\vskip 2pt}
\hline
\hline \noalign{\vskip 2pt}
XCLIP \cite{ma2022x} & \multicolumn{1}{c}{50.0} & 80.0 & \multicolumn{1}{c|}{8.8} & \multicolumn{1}{c}{64.8} & 91.1 & 3.0 \\ 
Simple CL & \multicolumn{1}{c}{50.2} & {80.1} & \multicolumn{1}{c|}{8.7} & \multicolumn{1}{c}{68.0} & {91.0} & {2.8} \\ 
Discernable  & \multicolumn{1}{c}{50.0} & 80.0 & \multicolumn{1}{c|}{8.7} & \multicolumn{1}{c}{67.9} & 92.2 & 3.0 \\ 
Adaptive  & \multicolumn{1}{c}{\textbf{50.7}} & {\textbf{80.3}} & \multicolumn{1}{c|}{\textbf{8.4}} & \multicolumn{1}{c}{\textbf{70.2}} & \textbf{94.0} & \textbf{2.3} \\ \noalign{\vskip 2pt}
\hline
\end{tabular}%
}
\vspace{-5pt}
\caption{Comparison of Video-Text Retrieval performance on MSVD \cite{MSVD} with batch size $30$.}
\label{tab:xclip_scores}
\vspace{-15pt}
\end{table}


\noindent
\textbf{QD-DETR \cite{qddetr}.} The findings for QD-DETR on the QVHighlights test set are outlined in Table \ref{tab:moment_scores} and it corroborates the observations made with Moment-DETR. Here, we consider the non-pretrained version of QD-DETR model as the baseline, which works with only video and text data. We use the same negatives used for the Moment-DETR model, initially incorporating them in the simple contrastive loss setting. This results in an average mAP score of $40.58$ which is a $1.8\%$ improvement from the baseline. For the most discernable setting, even though there is a minute decrease in average mAP, other metrics show improvement. This again supports the idea of introducing a mechanism for adaptive importance estimation of different parts of the sentence. Consequently, our attention-based adaptive importance mechanism provides an average mAP score of $40.66$, which is a $2.0\%$ improvement over the baseline. 

\vspace{-7pt}
\begin{table*}[!t]
\vspace{-5pt}
\begin{tabular}{c|cc|ccc|c|cc|ccc}
\hline \noalign{\vskip 2pt}
 & \multicolumn{5}{c|}{Moment-DETR \cite{qvh}} & & \multicolumn{5}{c}{QD-DETR \cite{qddetr}}\\ \cline{2-6} \cline{8-12}
 & \multicolumn{2}{c|}{R1} & \multicolumn{3}{c|}{mAP} & & \multicolumn{2}{c|}{R1} & \multicolumn{3}{c}{mAP}\\ 
 & \multicolumn{1}{c}{@0.5} & @0.7 & \multicolumn{1}{c}{@0.5} & \multicolumn{1}{c}{@0.75} & Avg & & \multicolumn{1}{c}{@0.5} & @0.7 & \multicolumn{1}{c}{@0.5} & \multicolumn{1}{c}{@0.75} & Avg\\ \noalign{\vskip 2pt} \hline 
\hline \noalign{\vskip 2pt}
Baseline & \multicolumn{1}{c}{53.87} & 34.00 & \multicolumn{1}{c}{55.42} & \multicolumn{1}{c}{28.96} & 30.90 & & \multicolumn{1}{c}{61.61} & 45.61 & \multicolumn{1}{c}{62.07} & \multicolumn{1}{c}{41.16} & 40.92\\ 

Negated Verb Query  & \textbf{57.35} & 39.55 & \textbf{57.43} & 33.69 & 34.45 & & \multicolumn{1}{c}{\textbf{63.35}} & \textbf{47.87} & \multicolumn{1}{c}{\textbf{62.94}} & \multicolumn{1}{c}{\textbf{42.76}} & \textbf{41.83}  \\ 

Negated Adjective Query & 55.94 & 38.13 & 57.12 & 34.04 & 35.10 & & \multicolumn{1}{c}{61.74} & 46.45 & \multicolumn{1}{c}{62.09} & \multicolumn{1}{c}{40.92} & 41.10 \\

Object Changed Query & 55.94 & 40.39 & 57.39 & 35.51 & \textbf{35.76} & & \multicolumn{1}{c}{61.94} & 46.26 & \multicolumn{1}{c}{62.42} & \multicolumn{1}{c}{41.45} & 41.24\\ 

Negated Passive Query & 56.26 & \textbf{41.29} & 57.21 & \textbf{35.75} & 35.63 & & \multicolumn{1}{c}{62.32} & 47.03 & \multicolumn{1}{c}{61.91} & \multicolumn{1}{c}{42.59} & 41.62 \\ 

Changed Subject Query & 56.84 & 38.06 & 57.37 & 33.98 & 34.28  & & \multicolumn{1}{c}{63.10} & 46.39 & \multicolumn{1}{c}{62.53} & \multicolumn{1}{c}{41.14} & 40.88

\\ \noalign{\vskip 2pt}
\hline
\end{tabular}%
\caption{Comparison of performance with baseline while using individual hard negatives on moment retrieval models: Moment-DETR and QD-DETR. The reported scores are on the validation set of the QVHighlights dataset.}
\label{tab:moment_ablation}
\vspace{-8pt}
\end{table*}

\vspace{-6pt}
\subsubsection{Video-Text Retrieval}
\vspace{-2pt}

\noindent
\textbf{X-CLIP \cite{ma2022x}.} Table \ref{tab:xclip_scores} outlines the result of X-CLIP model on the test set of MSVD \cite{MSVD} dataset. We generate hard negatives by changing the verb and the subject for this dataset. It can be observed that these hard negatives generated using LLM provide a significant improvement of $3.2\%$ in R@1 score of the text retrieval task and a marginal improvement of $0.2\%$  R@1 score of the video retrieval task just from simple contrastive loss. Furthermore, using weakly supervised importance estimation of sentence components, the model yields a significant improvement of $5.4\%$, $2.9\%$, and $0.7$ in R@1, R@5, and Mean Rank respectively for Video-to-Text (V2T) retrieval task and $0.7\%$, $0.3\%$ and $0.4$ improvement of R@1, R@5 and Mean Rank for the Text-to-Video (T2V) retrieval task, over the baseline. 
The huge improvement in the V2T task can be attributed to the model's high sensitivity to the differences between sentences and its hard negative, since this property is especially pertinent in the V2T task, which evaluates all the texts in the dataset for their relevance to a single video.
This further supports the hypothesis that introducing the weakly supervised adaptive importance estimation mechanism with generated hard negatives contributes to improved performance.

\subsection{Ablation Study and Discussion}
\subsubsection{Impact of sample generation criteria}

The models in video-language joint learning tasks fail to attend to all information present in default queries equally. To address this limitation, we generate negative instances in moment retrieval by targeting different components of a sentence such as verb, adjective/adverb, object, subject, and also by changing it to passive voice, as mentioned in the preceding section. Furthermore, for the case of video retrieval, we incorporate negatives by changing the verb and subject. We perform ablation studies to present how each of these generated negatives affects the overall performance by using a default contrastive learning setting with a single type of negative and a single positive at a time. 

\noindent\textbf{Moment Retrieval.} Table \ref{tab:moment_ablation} illustrates the impact of individual hard negative on the outcomes of moment retrieval models on the QVHighlights validation set. For Moment-DETR \cite{qvh}, every type of negative improves the performance compared to the baseline. This suggests that generating negative samples using any criterion has notable potential for improving the performance of video-text tasks. We also observe that the object-changed negative queries result in the best overall average mAP. This suggests that the baseline model provides comparatively low attention to objects present in queries. The table also presents the results for the QD-DETR \cite{qddetr} model where improvement in performance for each of the different negatives can be observed as well. Here we observe that the most improvement comes by utilizing negated verbs. This highlights that the baseline model has a limited capability in correlating query verbs with the videos in moment retrieval task.
 
\begin{table}[t]
\resizebox{\columnwidth}{!}{%
\small
\begin{tabular}{c|ccc|ccc}
\hline \noalign{\vskip 1pt}
 & \multicolumn{3}{c|}{T2V} & \multicolumn{3}{c}{V2T} \\ 
\textbf{} & \multicolumn{1}{c}{R@1$\uparrow$} & R@5$\uparrow$ & MnR$\downarrow$ & \multicolumn{1}{c}{R@1$\uparrow$} & \multicolumn{1}{c}{R@5$\uparrow$} & MnR$\downarrow$ \\ \noalign{\vskip 2pt}
\hline
\hline \noalign{\vskip 2pt}
XCLIP \cite{ma2022x}  & \multicolumn{1}{c}{50.0} & 80.0 & \multicolumn{1}{c|}{8.8} & \multicolumn{1}{c}{64.8} & 91.1 & 3.0 \\ 

Verb & \multicolumn{1}{c}{49.9} & \textbf{80.1} & \multicolumn{1}{c|}{8.7} & \multicolumn{1}{c}{\textbf{66.8}} & \textbf{93.2} & \textbf{2.9} \\

Subject & \multicolumn{1}{c}{\textbf{50.2}} & 80.0 & \multicolumn{1}{c|}{\textbf{8.6}} & \multicolumn{1}{c}{\textbf{66.8}} & 92.5 & \textbf{2.9}
\\ \noalign{\vskip 2pt}
\hline
\end{tabular}%
}
\caption{Contribution of different hard negatives with X-CLIP baseline on MSVD \cite{MSVD} with batch size $30$. `Verb': changing verb, `Subject': changing subject i.e. they represent the part of the caption that is modified to form a new caption.}
\label{tab:xclip_ablations}
\vspace{-10pt}
\end{table}

\noindent\textbf{Video Retrieval.} The results on the test set of the MSVD dataset, using individual negatives with X-CLIP \cite{ma2022x} as the baseline, are presented in Table \ref{tab:xclip_ablations}. Compared to baseline scores, notable improvement can be observed in the V2T metrics, with marginal improvements in T2V metrics as well. The results suggest that the baseline model in video-text retrieval has a limited capability in correlating both,  query verbs and subjects with videos.

 The increase in performance observed through these experiments substantiates our objective of generating automated hard negative samples utilizing LLMs by targeting specific parts of sentences in the text modality.

\begin{figure*}[t]
  \centering 
 \hspace*{-0.5cm}   \includegraphics[scale=0.188]{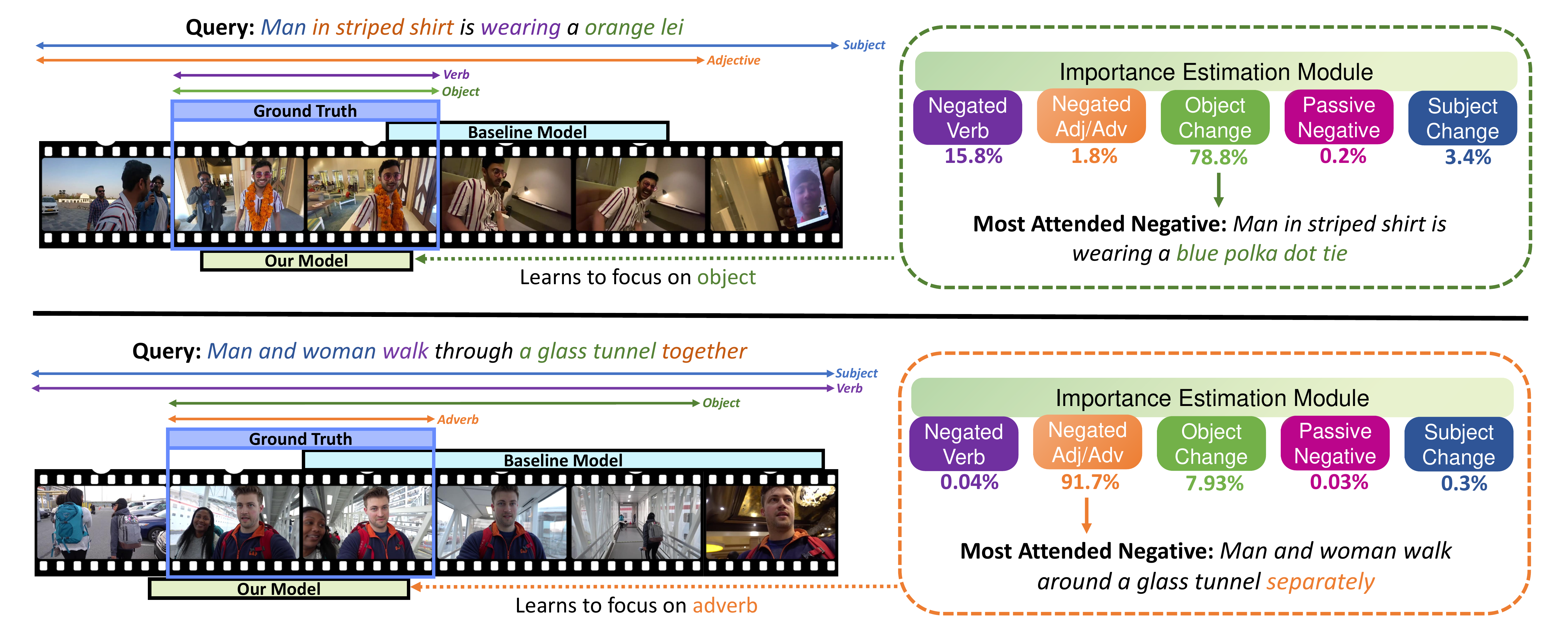}
   \caption{Examples of moment retrieval from baseline and our proposed version of Moment-DETR. The first example portrays a scenario where the original model struggled with the object \emph{orange lei} in the query which is remedied by our importance estimation module giving more weight to object changed negative which an LLM generated by changing the object to \emph{blue polka dot tie}. The second example provides a similar scenario where the model struggled with attending the adverb \emph{together} but is then remedied by our mechanism providing more weight to the adverb changed negative which an LLM generated by changing the adverb to its antonym \emph{separately}. In both cases, our mechanism learned to prioritize negative queries that address deficiencies in parts of the sentence where the baseline model struggled.}
   \label{fig:result}
   \vspace{-10pt}
\end{figure*}

\vspace{-10pt}
\subsubsection{Effect of Contrastive training}
\begin{figure}
 \centering
  \includegraphics[scale=0.45
]{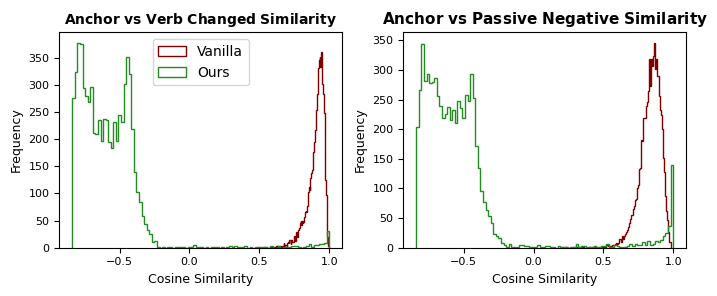}
  \vspace{-8pt}
   \caption{Distribution of similarity of sentence representation between ``Anchor and Verb-changed Negatives", and ``Anchor and Passive Negatives". Before applying our method, there were difficulties in discerning different types of texts, indicated by the high similarity of opposite texts. With our method, models effectively push negative text embeddings further from anchors, demonstrated by the low similarity of opposing texts.}
   \label{fig:distPlot}
   \vspace{-17pt}
\end{figure}
The main objective of using contrastive loss to incorporate the generated positives and negatives is to force the model to perceive the distinction among various words for a specific part of the sentence. We can infer from Figure \ref{fig:distPlot}, that the base model puts the anchor text and negative texts very close in the embedding space shown by their higher similarity of sentence representation. The proximity in embedding space often makes the model misjudge between opposite samples. Whereas, after the inclusion of the generated samples using contrastive loss, the model learns to push the negative samples further away from the anchor in embedding space. This is indicated by the very low similarity score between these opposite types of texts after inclusion.

\vspace{-7pt}
\subsubsection{Effect of Importance Estimation mechanism}
Our importance estimation module is based on the observation that the significance of a specific part of a sentence varies for different samples. A text has specific parts that are more relevant to the sample where the baseline model might not provide enough attention. To present that our proposed methodology is alleviating these deficiencies, we provide some samples comparing the outputs of the baseline models with proposed models in Figure \ref{fig:result}. In the first sample, the baseline model appears to be ignoring the object \emph{orange lei} in the query: ``\emph{Man in striped shirt is wearing a} orange lei". Our mechanism assigns the most importance to the respective object-changed hard negative sample, thus making the model attend to the object part of the query. The resultant model can successfully provide attention to the object resulting in better predictions. It is also noted that the second highest weighted negative is the negated verb query, which is relevant to the video and is also insignificantly attended by the baseline model.

\noindent While the baseline model provides less attention to the object in the first example, the second example illustrates an instance where the baseline model cannot discern the adverb \emph{together} in the query:  ``\emph{Man and woman walk through a glass tunnel} together". This further demonstrates the need for weakly supervised importance estimation for each video-text pair. We observe the proposed model assigning the highest weight to the respective negated adverb sample. Furthermore, in the video timeline, we observe that the baseline model also insufficiently recognizes the object \emph{glass tunnel} in the video and the corresponding object-changed negative is the second highest weighted negative sample. Consequently,  our proposed model can provide accurate predictions where all of the relevant parts of the query are represented. 




\section{Conclusion}
We present a novel framework for weakly supervised adaptive contrastive learning for multi-modal video-language tasks. Specifically, we mitigate the issue of models' deficiency in the perception of different sentence components by utilizing LLM-generated component-targeted negative samples. We additionally integrate our proposed adaptive importance estimation module to accommodate sample-wise variations in the significance of different types of sentence components. We evaluate our method across three different baselines and two different video-language joint learning tasks. In each task, our method outperforms the baseline considerably, validating the effectiveness of our proposed method.

{
    \small
    \bibliographystyle{ieeenat_fullname}
    \bibliography{main}
}

\clearpage
\setcounter{page}{1}
\maketitlesupplementary

\section{Attention Over Similarity Matrix of Text Data For Video-Text Retrieval}
\label{sec:AttentionBasedAggregation}
We compute the similarity score by incorporating the fine-grained representations of input texts to enhance the models' performance for video-text retrieval. It is a modified form of the one presented in the baseline \cite{ma2022x}. Considering the word-level representation for the texts $t$, $p$, and $n$, having dimension $\mathbb R^{tok_{i} \times dim}$, where $tok_{i}$ denotes the number of words/tokens in any text $i$; \{$t$, $p$\} forms a correct pair and \{$t$, $n$\} forms an incorrect pair. For any text pair $\{ a, b\}$, either correct or incorrect, a fine-grained similarity matrix, $\bS'_{a - b}\in\mathbb R^{tok_{a} \times tok_{b}}$, can be formulated as:

\begin{equation}
    \bS'_{a - b} = \ba (\bb)^{T}.
    \label{eqn:fine_grain}
\end{equation}
The resulting $\bS'_{a - b}$ is a similarity matrix containing similarity scores of every token of $a$ with every token of $b$. Simply averaging these scores to obtain the final similarity score would be ineffective for learning the relative importance of words and the alignment between them. Hence, we incorporate an attention-based similarity calculation by generating instance-level scores: $\bS'_{a}$ and $\bS'_{b}$, for $a$ and $b$ respectively.

Two stages of attention are applied to generate the instance-level similarity score, $\bS'_{a}$. From the first attention operation, we obtain $\bS'_{a}\in \mathbb R^{tok_b}$ by performing a weighted averaging of the similarity scores between each word in the word-level representation of $b$ with all of the words of the sentence embedding $a$ at a time, formulated as:
\begin{equation}
\bS'_{\ba} = \sum_{i=1}^{tok_a}\frac{\mathbf{e}^{ \bS'_{a - b(*, j)} / \tau }}{\sum_{j=1}^{tok_a} \mathbf{e}^{ \bS'_{\ba - \bb(j, *)} / \tau }}\bS'_{a - b(i, *)}.
\label{eqn:first_attention}
\end{equation}
The second attention operation performs weighted averaging on the output from the first attention operation, $\bS'_a$ to obtain the instance-level score for embedding $\ba$ ($\bS'_{a} \in \mathbb R$).
\begin{equation}
    \bS'_{a} = \sum_{i=1}^{tok_a}\frac{e^{ \bS'_{a(1, i)} / \tau }}{\sum_{j=1}^{tok_a} e^{ \bS'_{a(1, j)} \tau}}S_{a(1, i)}.
    \label{eqn:second_attention}
\end{equation}
Similarly, utilizing Equation \ref{eqn:first_attention} and \ref{eqn:second_attention} we generate the instance-level similarity, $\bS'_{b}$, for text $\bb$. 

After averaging these two instance-level similarity scores, $\bS'_{a}$ and $\bS'_{b}$, we get the final similarity score, $\bS_{a - b}$, of the sentence pair $\{a, b\}$,
\begin{equation}
    \bS_{a - b} = \left( \bS'_{a} + \bS'_{b} \right) / 2.
    \label{eqn:final_average}
\end{equation}
The $\texttt{sim}(\cdot, \cdot)$ function in Equation \ref{eqn:cons_act} returns this fine-grained similarity score, $\bS_{a - b}$, for video-text retrieval tasks.

Additionally, we use an auxiliary loss similar to the self-supervised cross-entropy loss in traditional video-text retrieval tasks. For this, we replace the original input texts to the model with the positive texts generated by the LLM with similar semantics. This loss further contributes to diversifying the video-language joint embedding space.

\section{Evaluation Protocols}\label{sec:eval_protocols}
\vspace{-5pt}
For moment retrieval, we utilize already established evaluation metrics. These are Recall@1 with $0.5$ and $0.7$ IoU thresholds, mean average precision (mAP) with $0.5$ and $0.75$ IoU thresholds along with the average mAP over a series of IoU thresholds (from $0.5$ to $0.95$ with an increment of $0.05$). On the other hand, for video-text retrieval we use the widely used retrieval metrics Recall at Rank K (R@K, higher is better), and Mean Recall (MnR, lower is better) for both video-to-text (V2T) and text-to-video (T2V) retrieval.

\section{LLM Prompting}\label{sec:llm_prompting}
\vspace{-5pt}
We use one-shot learning approach with the LLM to generate our additional text samples. Prompts used for generating a negative sample, \eg object-changed negative, and a positive sample are shown as Algorithm \ref{alg:LLM} and Algorithm \ref{alg:LLM2} respectively. We provide the instruction as the 
 \textbf{system} role of the LLM. Besides, we also pass an example of the operation in subsequent \textbf{user} and \textbf{assistant} roles to aid the LLM in understanding the task in the one-shot learning approach. Subsequently, the \textbf{user} queries the LLM with the anchor text as input, and the \textbf{assistant} outputs the sentence as required.

\vspace{-5pt}
\begin{algorithm}
\caption{Generation of Object-Changed Negative}
\label{alg:LLM}
\begin{algorithmic}[1]

\STATE \textbf{system} : Change the object of the sentence
\STATE \textbf{user} : A woman goes for a drive in a Greek island.
\STATE \textbf{assistant} : A woman goes for a drive in Sahara desert.
\STATE \textbf{user} : $\{input\}$
\STATE \textbf{assistant} : $\{output\}$ // output negative for $\{input\}$
\end{algorithmic}
\end{algorithm}
\vspace{-10pt}
\begin{algorithm}
\caption{Generation of Positive Sample}
\label{alg:LLM2}
\begin{algorithmic}[1]

\STATE \textbf{system} : Alter voice of the sentence
\STATE \textbf{user} : The chef cooks a meal.
\STATE \textbf{assistant} : A meal is being cooked by the chef.
\STATE \textbf{user} : $\{input\}$
\STATE \textbf{assistant} : $\{output\}$ // output positive for $\{input\}$
\end{algorithmic}
\end{algorithm}

\end{document}